\documentclass[letterpaper]{article} 
\usepackage{aaai24}  
\usepackage{times}  
\usepackage{helvet}  
\usepackage{courier}  
\usepackage[hyphens]{url}  
\usepackage{graphicx} 
\usepackage{natbib}  
\usepackage{caption} 
\usepackage{algorithm}
\usepackage{algorithmic}

\usepackage{multirow,booktabs}
\usepackage{amsmath}
\usepackage{amsfonts}

\usepackage{newfloat}
\usepackage{listings}
\DeclareCaptionStyle{ruled}{labelfont=normalfont,labelsep=colon,strut=off} 
\floatstyle{ruled}
\newfloat{listing}{tb}{lst}{}
\floatname{listing}{Listing}
\title{Spot the Error: \\Non-autoregressive Graphic Layout Generation with Wireframe Locator}
\author{
    Jieru Lin\textsuperscript{\rm 1}\thanks{Work was done during the first author's internship at Microsoft mentored by Danqing Huang.}, Danqing Huang\textsuperscript{\rm 2}, Tiejun Zhao\textsuperscript{\rm 1}, Dechen Zhan\textsuperscript{\rm 1}, Chin-Yew Lin\textsuperscript{\rm 2}
}
\affiliations{
    \textsuperscript{\rm 1}Harbin Institute of Technology, Harbin, China \\
    \textsuperscript{\rm 2}Microsoft Research\\

    hitjierulin@gmail.com, \{dahua, cyl\}@microsoft.com, \{tjzhao, dechen\}@hit.edu.cn
}

\begin{document}

\frenchspacing

\maketitle

\begin{abstract}
Layout generation is a critical step in graphic design to achieve meaningful compositions of elements. Most previous works view it as a sequence generation problem by concatenating element attribute tokens (i.e., category, size, position). So far the autoregressive approach (AR) has achieved promising results, but is still limited in global context modeling and suffers from error propagation since it can only attend to the previously generated tokens. Recent non-autoregressive attempts (NAR) have shown competitive results, which provides a wider context range and the flexibility to refine with iterative decoding. However, current works only use simple heuristics to recognize erroneous tokens for refinement which is inaccurate. This paper first conducts an in-depth analysis to better understand the difference between the AR and NAR framework. Furthermore, based on our observation that pixel space is more sensitive in capturing spatial patterns of graphic layouts (e.g., overlap, alignment), we propose a learning-based locator to detect erroneous tokens which takes the wireframe image rendered from the generated layout sequence as input. We show that it serves as a complementary modality to the element sequence in object space and contributes greatly to the overall performance. Experiments on two public datasets show that our approach outperforms both AR and NAR baselines. Extensive studies further prove the effectiveness of different modules with interesting findings. Our code will be available at https://github.com/ffffatgoose/SpotError.
\end{abstract}

\section{Introduction}
Layout generation refers to the arrangement of elements (i.e., size and position) on a canvas, which is essential for creating visually appealing graphic designs (e.g., articles, user interface). State-of-the-art systems~\cite{layoutvae_2019,vtn_2021,layoutgan_plus_2021,layouttransformer_2021} mostly view the task as a sequence generation problem where the sequence is composed of element attribute tokens (i.e., category, position, size). Majority of the works follow the autoregressive (AR) approach which generates one token at a time based on the previously output and have achieved promising results~\cite{layouttransformer_2021,Guo2021TheLG,yang2023Intermediate,weng2023graph}.
However, for layout generation, it is important to model relationships between elements as well as the global context.
The inherent causal attention in the AR approach exposes limitation in global context modeling and causes immutable dependency chain issue and error propagation~\cite{blt}.

\begin{figure}[!tb]
\centering
\includegraphics[scale=0.4]{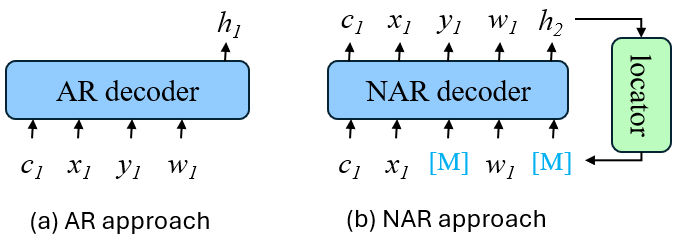}
\caption{Illustration of (a) AR and (b) iterative-based NAR approach in layout generation. The AR approach generates one token at a time conditioned on previously generated tokens. NAR generates all tokens simultaneously and uses a locator (usually heuristics) to detect erroneous tokens which will be masked and re-predict in the next decoding iteration.}
\label{fig:intro}
\end{figure}

\begin{figure*}[htb]
\centering
\includegraphics[scale=0.25]{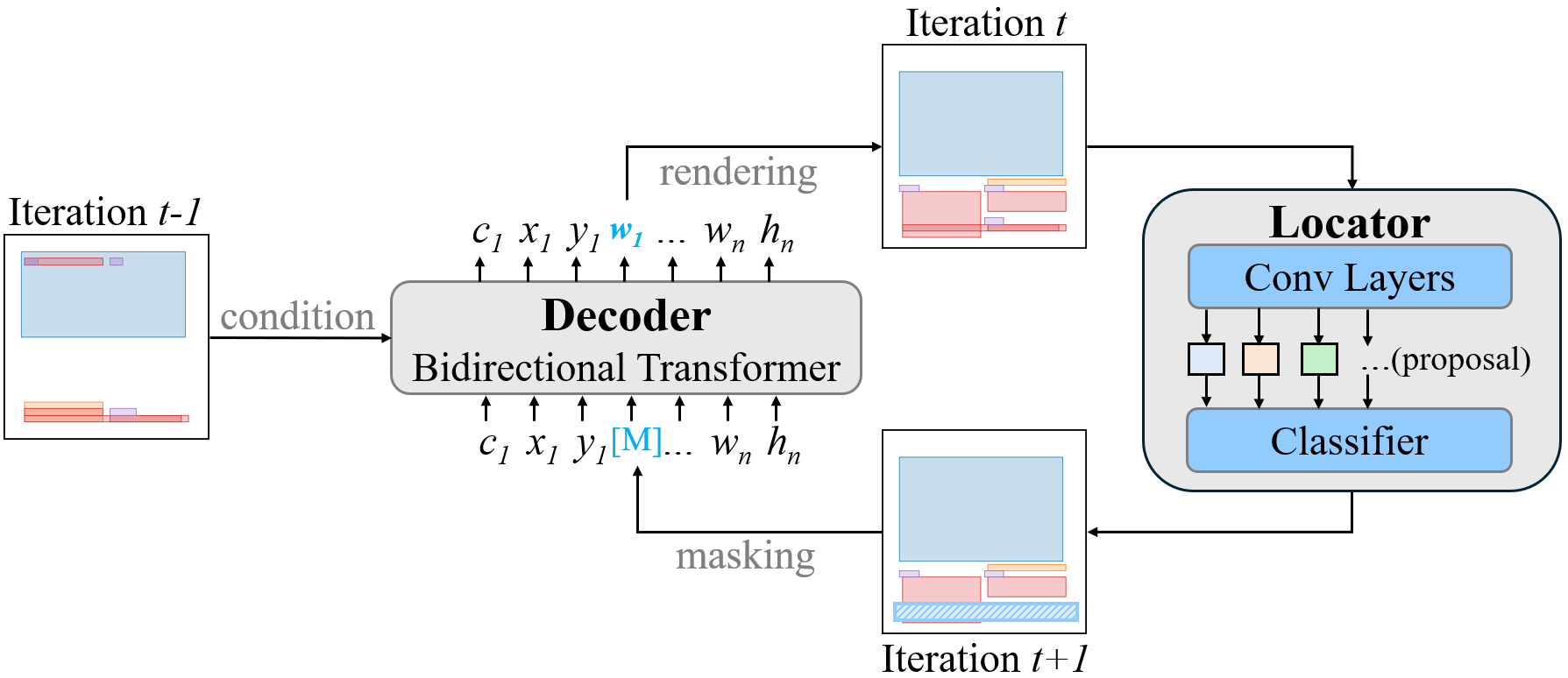}
\caption{Overview of our pipeline. Our model consists of a decoder and a locator. For each mask-predict iteration, the locator detects the erroneous attribute tokens to be masked and the decoder predict the masked tokens in a non-autoregressive approach.
}
\label{fig:pipeline}
\end{figure*}

Recently, there are some attempts of non-autoregressive (NAR) approaches~\cite{blt,zhang2023layoutdiffusion,layoutdm2023,decoupleDiffusion_2023} that generate the entire sequence simultaneously. It allows a more flexible modeling of bidirectional context and shows promising results in the task.
In this paper, we conduct an in-depth analysis to compare the AR and NAR approaches (as shown in Figure~\ref{fig:intro}) with two findings: (1) we show that NAR is more robust to different element orders in the sequence than AR; (2) similar to the word repetition issue in non-autoregressive machine translation (NMT)~\cite{nat2022distill}, NAR tends to position elements to similar regions that will cause overlap problem and results in inferior layouts.

Iterative decoding is an effective mechanism to alleviate this issue. In the recent state-of-the-art NAR system BLT~\cite{blt}, it uses a Transformer-based decoder~\cite{vaswani2017attention} and applies a heuristic strategy to choose the least confident tokens from the decoder's prediction as the erroneous ones. However, such simple heuristic is likely to be biased to the decoder's learned distribution and potentially propagate the mistakes, acting as noises to make a negative impact on subsequent iterations. 

This paper proposes a learning-based locator module to explicitly distinguish the incorrect tokens. 
Using a toy experiment of comparing different layout representations, we observe that pixel space is more sensitive in capturing spatial patterns of layouts (e.g., overlap, alignment). Therefore, our locator receives the input of wireframe image rendered from the decoder's generated sequence in the previous iteration, and then detects the erroneous element and classify the corresponding attribute tokens if they should be masked or not. The wireframe serves as a complementary modality to the layout sequence in object space which we will show is effective. 
To train the locator, it requires the annotation of erroneous attributes in a noisy layout (e.g., the width of an element is inaccurate and should be masked). One possible solution is that we make random noise on the real layouts without human labeling. However, our initial experiment shows minor improvement when applying the locator to the refinement process. In order to focus more on the decoder's error types and align with its distribution, we propose a novel automatic data construction pipeline. Specifically, we use the decoder's generated layout to retrieve the most similar real layout as ground truth and apply Hungarian matching between the two sets of elements. For the matched element pairs that with significant attribute value differences, we annotate the corresponding tokens as the targets to be masked.

We evaluate our approach on two public datasets related to graphic design: RICO~\cite{deka2017rico} for mobile app UIs and PubLayNet~\cite{zhong2019publaynet} for scientific articles. Experiment results show that our approach outperforms current AR and NAR baselines in terms of both quantitative and qualitative evaluations. Furthermore, we conduct extensive analysis to better understand what our model has learned.

To summarize, the contributions of this paper include:
\begin{itemize}
    \item We conduct an in-depth analysis to compare between AR and NAR models in layout generation. Though NAR is robust to different input element orders, it exposes the issue of repetitive token generation.
    \item We investigate the use of different layout representations (i.e., object, pixel) with a toy experiment. We observe that pixel space is more sensitive in capturing spatial patterns than object space, which motivates us to incorporate the image modality into the modeling process.
    \item To improve the iteration-based NAR, this paper proposes a learning-based locator that takes a rendering wireframe as input and detects erroneous tokens more accurately.
\end{itemize}

\section{Related Works}

\subsection{Layout Generation}
Layout generation is an important task in graphic design intelligence~\cite{huang2023survey,li2021towards} (e.g., layout representation learning~\cite{Xie2021CanvasEmb,feng2022rep}, layout reverse engineering ~\cite{hao2023reverse,shi2023reverse,Zhu2024grouping, huang2021visual}). 
Traditional works~\cite{hurst2009review,kumar2011bricolage,o2014learning,tabata19automatic} are mostly based on heuristics with constraint optimization, which usually ensure high-quality but limited outputs. Recently there are an increase in works based on deep neural networks.
LayoutGAN~\cite{layoutgan2019} applies GAN to synthesize the layout bounding box and proposes a differential wireframe rendering module to enable the training of discriminator, and LayoutGAN++ \cite{layoutgan_plus_2021} extends LayoutGAN with Transformer backbone.
LayoutVAE \cite{layoutvae_2019} trains two VAEs separately, one to predict the element categories and the other to generate layouts given the category condition. Several methods also follow the VAE-based generative framework~\cite{ndn_2020,jiang_coarse_fine_2022}.
Recent works \cite{layouttransformer_2021,vtn_2021}  build generative backbone based on Transformer \cite{vaswani2017attention} to model long-distance dependency and perform better. 

BLT~\cite{blt} is the first recent attempt of using non-autoregressive in layout generation. The decoder is trained using BERT-like strategy to predict some randomly masked tokens. During inference, it uses iterative decoding that tokens with the least confident score are masked and further refinement. Diffusion-based models also follow the NAR framework~\cite{layoutdm2023, zhang2023layoutdiffusion,decoupleDiffusion_2023} which progressively infers a complete layout from a noisy status with discrete diffusion process. Despite different modeling approaches, the above mentioned methods use element sequence to represent a layout.
This paper observes that layouts represented in pixel space offers a different aspect of features and hence we propose to combine the object space and pixel space in the NAR framework for better performance.

There are also some content-aware generation methods \cite{zheng_magazine_2019,wang_aesthetic_2022,cao_geometry_2022,zhou_composition-aware_2022,li_harmonious_2022,vaddamanu_harmonized_2022} that further considers the element content into modeling, which we will leave for future exploration.

\subsection{Non-autoregressive Sequence Generation}
Comared to AR~\cite{huang2018neural,zhou2019towards,huang-etal-2018-using}, NAR is more efficient in inference and has been applied in various tasks such as neural machine translation, automatic speech recognition and text to speech. Here we mainly focus on the related works of iteration-based non-autoregressive machine translation (NMT)~\cite{nmtRewrite2021,cmlm2019,nat2022distill,saharia2020_non}. These iterative decoding methods using different masking strategies, including heuristics based and and learning-based. Our locator is inspired from the latter approach.
For a more complete review, please refer to the survey~\cite{natSurvey2022}. 

\section{Preliminary Study}

\subsection{Task Formulation}
Layout generation can be viewed as a sequence generation:
$$
\mathbf{s} = ([bos], c_1, x_1, y_1, w_1, h_1, ..., c_n, x_n, y_n, w_n, h_n, [eos])
$$
where $c_i$ is the category label of the $i$-th element in the layout (e.g., \texttt{title}, \texttt{text}, \texttt{figure}), $x_i, y_i, w_i, h_i$ represent the position and size which are converted to discrete tokens. $[bos], [eos]$ are special tokens for beginning and end. The total sequence length is $5n+2$. 

Autoregressive (AR) generation predicts the token one at a time, conditioning on the previous generated sequence: 
\begin{equation}
    p(\mathbf{s}) = \prod_{i=2}^{5n+2}p(s_{i}|s_{1:i-1}) 
\end{equation}
while in the non-autoregressive setting,  the model predicts one token with bidirectional attention and predicts the whole sequence simultaneously:
\begin{equation}
    p(\mathbf{s}) =  \prod_{i=2}^{5n+2}p(s_{i}|s_{1:i-1}, s_{i+1:5n+2}) 
\end{equation}

\subsection{Non-Autoregressive Decoding Analysis}
To better understand how different decoding methods (i.e., AR and NAR) behave in the layout synthesis process, this paper conducts an in-depth analysis with some interesting findings. In the following analysis,  we use the state-of-the-art AR model LayoutTransformer as well as the  the NAR model BLT on the scientific article dataset PubLaynet.

\begin{figure}[!tb]
\centering
\includegraphics[scale=0.40]{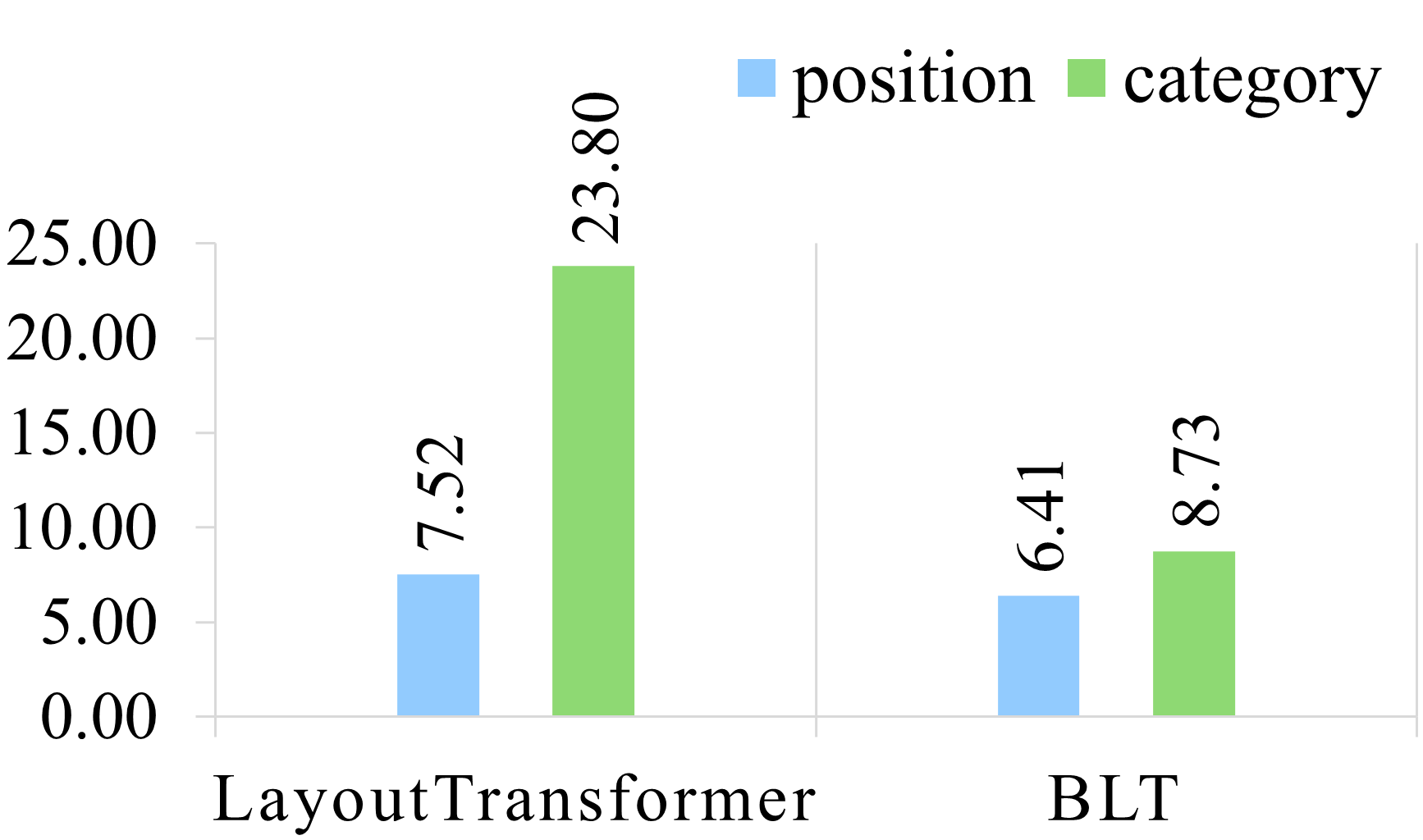}
\caption{Comparison of AR (LayoutTransformer) and NAR (BLT) approaches using different input element orders (i.e., position, category). Smaller overlap degree indicates better performance.}
\label{fig:element_order}
\end{figure}

\paragraph{Finding 1: NAR is robust to input element order.} For sequence generation, the input order is an important factor. We compare the AR and NAR models with different element orders used in previous works~\cite{layouttransformer_2021,blt}: (1) position, where elements are sorted using the top-left coordinates. While most previous works follow this setting, it actually causes  information leak of the ground truth data during inference since they use absolution positions in real layouts to determine the order;  (2) category, where input elements are fed per category (e.g., generate all the \texttt{paragraph} elements first, then followed by \texttt{table}). Figure~\ref{fig:element_order} shows the performances in terms of the \texttt{Overlap} metric, which calculates the average overlap degree between elements in a layout and is widely used for evaluation:
\begin{equation}
    \mathrm{Overlap} = \frac{1}{D} \sum_{d=1}^{D} \Big[ \sum_{i=1}^{N} \sum_{\substack{j \neq i}} \frac{e_i \cap e_j}{e_i} \Big]
\end{equation}
where $D$ denotes the number of layouts, $N$ is the element number of a layout, $e_i$ indicates the $i$-th element in a layout, $e_i \cap e_j$ means the overlap area between $e_i$ and $e_j$.

Based on the general assumption that elements in a layout should not overlap excessively, smaller value indicates better quality. From the figure, we can see that the AR model is more sensitive to orders as its performance drops significantly from the `category' setting  to the `position' one. Being compared, the NAR model is relatively robust to different input orders. 

\paragraph{Finding 2: NAR exposes the issue of repetitive generation.} Compared to AR with causal masked attention over previously generated tokens, NAR applies a full attention over all tokens on a wider range of contexts, which is difficult for the model to distinguish different tokens by solely relying on positional encodings. Thus it is more likely to generate repetitive tokens~\cite{nat2022distill}. Here we try to quantify this intrinsic error using the aforementioned \texttt{Overlap} metric. Higher overlap degree indicates that the model is more likely to generate tokens in a similar region. Table~\ref{tab:repetition} shows the statistics considering overlaps between (1) elements within the same category and (2) all elements. We can see that the overlap problem in NAR is more serious than AR (26.89 compared to 23.80 for all elements), and nearly two times the degree under the `same category' setting (15.84 compared to 8.67). Furthermore, we show that iterative refinement is an effective mechanism to alleviate the issue as the overlap (all) decreases from 26.89 (iter 1) to 10.23 (iter 10).

\begin{table}[!tbp]
    \centering
    \resizebox{0.9\columnwidth}{!}{
    \begin{tabular}{lccc}
    \toprule
    \multirow{2}{*}{Models} & \multirow{2}{*}{Iter} & \multicolumn{2}{c}{Overlap} \\
    \cmidrule(lr){3-4}
    & & same cat. & All \\
    \midrule
    LayoutTransformer (AR) & / & 8.67 & 23.80 \\
    \midrule
    BLT (NAR) & 1 & 15.84 & 26.89  \\
    BLT  & 5 & 5.20 & 13.43  \\
    BLT  & 10 & 3.23 & 10.23 \\
    \bottomrule
    
    \end{tabular}
    }
    \caption{NAR generates a larger percentage of repetitive tokens than the AR approach. The issue is alleviated with iterations of refinement.}
    \label{tab:repetition}
\end{table}

In this paper, we focus on the iteration-based NAR generation approach and propose a simple but effective locator module to improve the iterative refinement process.

\subsection{Layout Representation: Object v.s. Pixel} 

Previous layout generation methods are far more based on primitive elements with attributes to form a sequence. Meanwhile, layouts can also be represented by a wireframe image which is rendered by the element attributes, which provides a more direct visual perception~\cite{layoutgan2019}.
We argue that both representations are informative for capturing important features in different aspects. 

To verify our claim, we design a toy experiment of detecting the erroneous element attributes in a noisy layout using the two different representations respectively. For training and evaluation, we sample elements in real layouts as good layouts, and add random noise on some attributes $x,y, w, h$ as the bad layouts. In the object space given a sequence of element attributes, we adopt a Transformer with a classification layer on the top to predict good/bad for each token. In the pixel space we use the convolutional-based Faster-RCNN~\cite{fasterrcnn2015} to detect the inferior elements and classify each of its attributes. Table~\ref{tab:analysis-layoutMask} shows the classification results. As we can see, when the noise range is large (e.g., 0.5), models in both spaces can detect accurately with F1 score over 96\%. If the noise is decreased to a smaller range (e.g., 0.1) where the noisy layouts are less visually different from the real ones, models in pixel space performs significantly better. This indicates that pixel representation is more sensitive in capturing fine-grained spatial patterns such as minor misalignment and occlusion.

\begin{table}[htbp]
    \centering
    \resizebox{0.95\columnwidth}{!}{
    \begin{tabular}{lcccc}
    \toprule
    Repre. & Noise range & Precision & Recall & F1 Score \\
    \cmidrule(lr){1-5}
    Object & 0.1 & 84.80 & 79.70 & 82.17 \\
    Pixel & 0.1 & 92.19 & 88.78 & 90.45  \\
    \midrule
    Object & 0.2 & 96.36 & 93.62 & 94.97 \\
    Pixel & 0.2 & 96.56 & 95.73 & 96.14  \\
    \midrule
    Object & 0.5 & 99.16 & 93.08 & 96.02 \\
    Pixel & 0.5 & 97.35 & 95.58 & 96.46 \\
    \bottomrule
    \end{tabular}
    }
    \caption{Classification results of our toy experiment to compare different layout representations (i.e., object and pixel space) using random noise data.} 
    \label{tab:analysis-layoutMask}
\end{table}

\section{Our Approach}

We generate graphic layouts using the non-autoregressive approach. Our model consists of a \textit{decoder} to generate the tokens and a \textit{locator} to recognize the erroneous tokens which will be revised by the decoder iteratively. Figure~\ref{fig:pipeline} shows the pipeline overview.

\subsection{Decoder}
Following BLT~\cite{blt}, we use a multi-layer transformer decoder to predict the layout sequence in parallel. Each token in the sequence is represented by the sum of its attribute embedding (i.e., category, position or size) and three additional position encodings $\gamma$ to better distinguish different tokens, (1) the token index in a sentence; (2) the element-level index where tokens belonging to the same element will have the same value. (3) the number of elements where all tokens share the same value. 
\begin{equation}
\begin{aligned}
  PE(i) & = \gamma_{1} (i ) + \gamma_{2} (\lfloor i / 5 \rfloor ) + \gamma_{3} (n) \\
\end{aligned}
\end{equation}
where $i \in 5n+2$.

During training, we randomly mask some tokens in the layout sequence $\mathbf{s}$ similar to BERT~\cite{devlin2018bert} and compute the loss on these mask positions $M$ to minimize the negative log-likelihood:
\begin{equation}
    \mathbf{\mathcal{L}_{mask} }= - \mathbb{E}_{\mathbf{s} \in \mathcal{D}} \Big[ \sum_{i \in M } \log p(s_i | s^M)\Big]
\end{equation}

In inference, we replace all the attributes which are not given with \texttt{[MASK]} token (in conditional generation, the element categories are given) to initialize the input sequence and the decoder is then used to predict the masked tokens.

\paragraph{Wireframe Conditioning.} Given the wireframe image $I$ encoded with a convoluational network~\cite{lecun1998gradient} $\varphi$, we apply the cross-attention of the Transformer-based decoder to the image, which provide the model a spatial context for learning to attend to relevant regions:
  
\begin{equation}  
  \text{Attention}(\mathbf{s}, I) = \text{softmax}\left(\frac{\phi(\mathbf{s}) \varphi(I)^T}{\sqrt{d_k}}\right)
  \label{equ:crossAtten}
\end{equation}  
where $\phi(\mathbf{s})$ is the query matrix derived from the transformer decoder, $\varphi(I)$ is the key matrix from the convolutional network, respectively, and $d_k$ is the dimension of the key $\varphi(I)$.

\subsection{Locator}
\label{sec:locator}

In the iteration-based NAR models, the dominant approach to identify the erroneous tokens for masking and re-predict is to employ heuristic rules to roughly choose the least confident tokens from decoder output. However, relying on the decoder`s confidence score is likely to cause bias to its learned distribution, and the heuristic rules might not be sufficient to cover a wide range of cases.

This paper proposes to utilize a learning-based locator module to select the tokens. The locator consists of three components: a backbone network, a mask classifier, and a bounding box predictor. Given the wireframe input $I$, the backbone network extracts features $F_{feat}$ from $I$, the mask classifier binary classifies each of the attributes ($x, y, w, h$) to determine if it should be masked or not:
\begin{align}  
    F_{feat} &= \mathrm{Conv}(I), \\  
    C_{mask} &= \mathbf{W_c}\mathrm{RPN}(F_{feat}) + \mathbf{b_c}
\end{align}
where $\mathbf{W_c}$ and $\mathbf{b_c}$ are the parameters of mask classifier.

To train the locator, there is no direct annotated data of noisy layout with targeted erroneous tokens. Here we introduce a novel algorithm to construct the training data, which is described in Algorithm~\ref{alg:locator}. Specifically, we collect the generated noisy layouts from the trained decoder in different iterations. For each layout, we retrieve the ground truth layouts with the same element categories and conduct a bipartie matching between elements in the generated layout $\{e_i^{g}\}$ and the real one $\{e_j^{r}\}$. The matching function $d$ considers the category constraint (element categories must be the same) and the spatial relationships (i.e., the bounding box overlap ratio, position distance, and the size disparities):

\begin{equation}  
\begin{aligned}  
d(e_i^g, e_j^r)  &= \alpha_1 \mathbb{1}_{c_i \ne c_j} + \alpha_2 \mathrm{IoU}(e_i^g, e_j^r)  \\  
                &+ \alpha_3 f_{pos}(e_i^g, e_j^r) + \alpha_4 f_{area}(e_i^g, e_j^r), \\
\end{aligned}
\label{equ:costMetric}
\end{equation}

where $f_{pos}$ and $f_{area}$ are the functions which compute the position and area differences. 

After matching, we choose elements whose overlap with the ground truth smaller than a threshold as the mistakes and annotate its corresponding attributes to be masked.

\begin{algorithm}[!tbp]
    \caption{Training data construction for locator.}
    \label{alg:locator}
    \begin{algorithmic}[1]
        \REQUIRE Generated layouts $L^g$ with $n$ elements, ground truth layout ${L^{real}}$ with the same categories $(c_0, c_1, \ldots, c_n)$, threshold $\delta$
        \STATE erroneous attribute set to be masked $M[k]=\{\}$ for $k$ in attributes $\{x, y, w, h\}$
        \STATE Set matching $\{e_i^{g}\}$ and $\{e_j^{real}\}$ with distance function $d$
        \FOR{$i = 1, \cdots, n$}
            \IF {$\mathrm{abs}(e_i^{g}[k]-e_j^{real}[k]) > \delta$}
                \STATE $M[k]$ = $M[k]$ + $e_i^g[k]$
            \ENDIF
        \ENDFOR
    \end{algorithmic}
\end{algorithm}

\begin{table*}[ht]
    \centering
    \setlength{\tabcolsep}{1mm}
    \resizebox{2.05\columnwidth}{!}{
    \begin{tabular}{lcccccccccc}
    \toprule
        \multirow{2}{*}{\textbf{Models}} & \multicolumn{5}{c}{\textbf{PubLaynet}} & \multicolumn{5}{c}{\textbf{Rico}}\\
        \cmidrule(lr){2-6} \cmidrule(lr){7-11}
         & SeqFID$\downarrow$ & PixelFID$\downarrow$ & Max IoU$\uparrow$ & Overlap$\downarrow$ & Alignment$\downarrow$ & SeqFID$\downarrow$ & PixelFID$\downarrow$ & Max IoU↑ & Overlap$\downarrow$ & Alignment$\downarrow$ \\
         \cmidrule(lr){1-11}
         Layout VAE & 27.18 & 158.72 & 0.2473 & 45.81 & 0.6617 & 87.61 & 12.52 & 0.3678 & 63.20 & 0.8709 \\
         LayoutTransformer & 10.20 & 76.96 & 0.3465 & 23.80 & 0.0912 & 44.77 & 16.31 & 0.4165 & \textbf{61.04} & \textbf{0.0731} \\
         VTN  & 8.60 & 78.76 & 0.3122 & 26.51 & 0.2196 & 18.49 & 28.34 & 0.4332 & 63.22 & 0.4411 \\
         Layout GAN++ & 9.52 & 53.71 & \textbf{0.4827} & 14.50 & 0.2019 & \textbf{13.49} & 6.78 & 0.5038 & 65.14 &  0.3087 \\
         BLT & 7.88 & 4.60 & 0.3865 & 8.73 & 0.0765 & 32.67 & 3.80 & 0.5669 & 72.64 & 0.2361 \\
         BLT w/o HSP & \textbf{6.00} & 3.82 & 0.3985 & 9.21 & 0.0988 & 32.06 & 3.36 & 0.5773 & 68.66 & 0.2152 \\
         \cmidrule(lr){1-11}
         \textbf{Ours (dec-only)} & 6.30 & 2.52 & 0.3959 & 8.64 & 0.0740 & 31.21 & \textbf{2.65} & \textbf{0.5807} & 66.33 & 0.1995 \\
         \textbf{Ours (w/ locator)} & 9.99 & \textbf{2.33} & 0.2962 & \textbf{7.57} & \textbf{0.0634} & 27.04 & 2.82 & 0.5693 & 62.25 & 0.2075 \\
         \cmidrule(lr){1-11}
         Real data& 4.91 & 0.02 & 0.5300 & 0.22 & 0.0400 & 4.26 & 1.16 & 0.6600 & 48.43 & 0.2000 \\
        \bottomrule
    \end{tabular}
    }
    \caption{Conditional generation results when given element category (C$\rightarrow$SP). The best results are in bold.}
    \label{tab:constrainC-SP}
\end{table*}

\begin{table*}[ht]
    \centering
    \setlength{\tabcolsep}{1mm}
    \resizebox{2.05\columnwidth}{!}{
    \begin{tabular}{lcccccccccc}
    \toprule
        \multirow{2}{*}{\textbf{Models}} & \multicolumn{5}{c}{\textbf{PubLaynet}} & \multicolumn{5}{c}{\textbf{Rico}}\\
        \cmidrule(lr){2-6} \cmidrule(lr){7-11}
         & SeqFID$\downarrow$ & PixelFID$\downarrow$ & Max IoU$\uparrow$ & Overlap$\downarrow$ & Alignment$\downarrow$ & SeqFID$\downarrow$ & PixelFID$\downarrow$ & Max IoU↑ & Overlap$\downarrow$ & Alignment$\downarrow$ \\
         \cmidrule(lr){1-11}
         LayoutVAE & 22.93 & 134.90 & 0.3295 & 45.59 & 0.6257 & 86.64 & 7.71 & 0.4124 & 63.79 & 0.4899 \\
         LayoutTransformer & 26.49 & 130.77 & 0.3131 & 36.04 & 0.3261 & 59.81 & 11.77 & 0.4262 & \textbf{59.42} & \textbf{0.1737} \\
         BLT & 2.39 & 3.04 & \textbf{0.5502} & 8.96 & 0.1404 & 20.07 & 2.26 & 0.5737 & 69.31 & 0.3941 \\
         BLT w/o HSP & 2.27 & 1.60 & 0.5493 & 8.47 & 0.1356 & 18.08 & 2.17 & 0.5800 & 64.73 & 0.2858 \\
         \cmidrule(lr){1-11}
         \textbf{Ours (dec-only)} & 2.09 & 1.51 & 0.5450 & \textbf{8.44} & \textbf{0.1275} & 19.42 & 1.91 & 0.5774 & 65.24 & 0.2647 \\
         \textbf{Ours (w/ locator)} & \textbf{2.04} & \textbf{1.44} & 0.5428 & 9.69 & 0.1280 & \textbf{17.88} & \textbf{1.88} & \textbf{0.5801} & 62.47 & 0.3070 \\
          
         \cmidrule(lr){1-11}
         Real data& 4.91 & 0.02 & 0.5300 & 0.22 & 0.0400 & 4.26 & 1.16 & 0.6600 & 48.43 & 0.2000 \\
        \bottomrule
    \end{tabular}
    }
    \caption{Conditional generation results given element category and size (CS$\rightarrow$P). The best results are in bold.}
    \label{tab:constrainCS-P}
\end{table*}

\section{Experiments}
\subsection{Dataset}

We experiment with two widely used public datasets in different design types: Rico \cite{deka2017rico,liu2018learning} and Publaynet \cite{zhong2019publaynet}. 
\textbf{Rico} is a UI/UX design dataset consisting of over 66k UI layouts collected from Android mobile apps. We filter out layouts with the number of elements in a layout more than 9 and use the most common 13 category in Rico following the previous work~\cite{layoutgan_plus_2021}. There are 20,606 layouts in total finally. We split train/val/test dataset at a rate of 0.85/0.05/0.10. 
\textbf{PubLaynet} contains over 360k layouts of scientific paper pages from PubMed Central. We use the official split train and test set. Similarly, we exclude layouts elements more than 9, which are in total 173,225 layouts.
For inference, we use the `category` input order mentioned in previous section for conditional generation.

\subsection{Evaluation Metrics}

There are 4 metrics commonly used to measure the generated layout quality:
\begin{itemize}
    \item \textbf{Maximum IoU.} Given the generated layouts and the references, this metric computes the intersection over the union of the two sets with a permutation to maximize the IoU as a similarity measurement.
    \item \textbf{Alignment.} Layout elements are usually aligned with each other to create an organized composition. Alignment calculates on average the minimum distance in the x- or y-axis between any element pairs in a layout.
    \item \textbf{Overlap.} It is assumed that elements should not overlap excessively. Overlap computes the average IoU of any two elements in a layout. Layouts with small overlap values are often considered to be high quality.
    \item \textbf{FID.} Compared to the above heuristic metrics, FID is a sample-based metric for image generation~\cite{heusel2017gans} and has been adopted in layout generation. It pretrains a feature network to classify real or fake layouts which is then used to extract features of two data sets and calculate the Fréchet distance. Here we use two pre-trained FID, namely SeqFID~\cite{layoutgan_plus_2021} and PixelFID~\cite{yang2023Intermediate} using different layout representations (object and pixel respectively). Please refer to \citet{yang2023Intermediate} for a more comprehensive comparison between the two metrics.
\end{itemize}

\subsection{Baselines}
We consider the following public-available works as baselines, including the \textbf{autoregressive} models: \textbf{LayoutVAE}~\cite{layoutvae_2019} takes the latent code and category labels (optional) as input and generates the element bounding boxes in an autoregressive manner.  
\textbf{VTN}~\cite{vtn_2021} uses transformer layer to enhance the performance of VAE as well as increase the number of elements in layouts.
\textbf{LayoutGAN++}~\cite{layoutgan_plus_2021} improves LayoutGAN with Transformer backbone and applies several beautification post-process for alignment and non-overlap. 
\textbf{LayoutTransformer}~\cite{layouttransformer_2021} autoregressively generates a sequence of element tokens. And we also compare with the state-of-the-art \textbf{non-autoregressive} model BLT~\cite{blt}. Specifically, they apply a heuristic strategy \textbf{HSP} for iterative decoding: attributes are grouped into category, size and position, and tokens in different groups are predicted in a pre-defined order. In our experiment, we observe that such strategy is not always effective so we also show the result \textbf{BLT w/o HSP}.

\begin{figure*}[htb]
\centering
\includegraphics[scale=0.4]{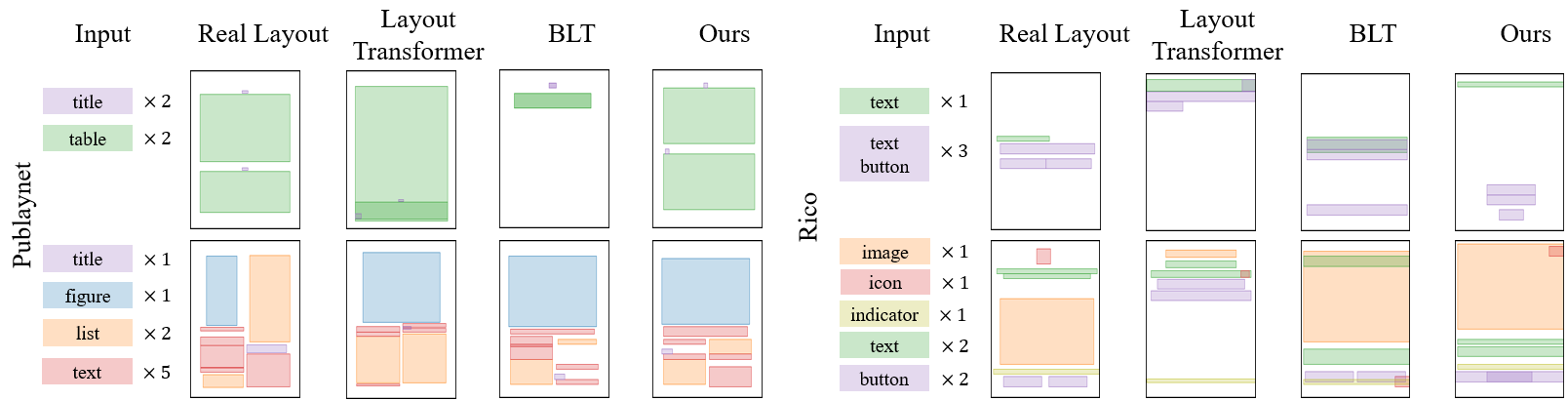}
\caption{Qualitative results on Publaynet and Rico.
}
\label{fig:example}
\end{figure*}

\subsection{Main Results}
Following BLT, we consider two settings of conditional generation: (1) Condition on Category (C$\rightarrow$SP), which only element categories are given as input and the model needs to predict the element size and position; (2) Condition on Category + Size (CS$\rightarrow$P), which the element category and size are specified and the model predicts only the positions. We show the performances in Table~\ref{tab:constrainC-SP} and Table~\ref{tab:constrainCS-P} respectively\footnote{Due to space limit, we show the unconditional task in Suppl.}.

As we can see, the NAR baseline BLT can already achieve similar performances as the AR approaches, and even significantly better on PubLaynet under both C$\rightarrow$SP and C$\rightarrow$SP settings. For our approach, only using the wireframe-conditioned decoder (dec-only) achieves promising results (e.g., PixelFID 2.52 compared to 4.60 in BLT on PubLaynet under C$\rightarrow$SP setting). Coupled with our learning-based locator, we reach the best results in most of the metrics. Please note that for the two FID evaluators, we see similar trends in relative comparison between different systems, but PixelFID is proved to be more robust and unbiased to different decoder architecture ~\cite{yang2023Intermediate}.

\begin{figure}[htb]
\centering
\includegraphics[scale=0.18]{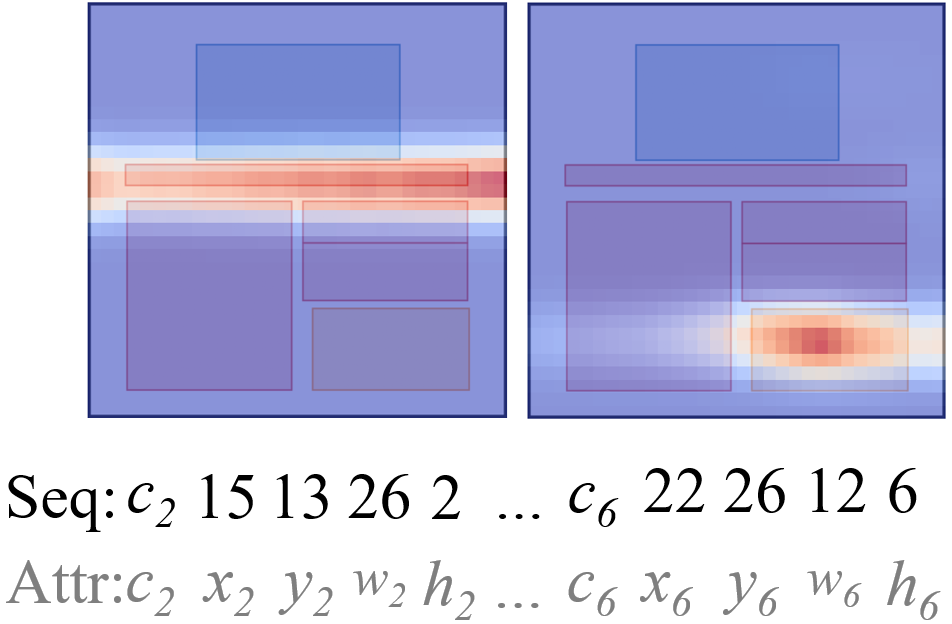}
\caption{Decoder-Wireframe cross-attn visualization.}
\label{fig:CrossAttenHeatMap}
\end{figure}

\subsection{Qualitative Results}
Here we show some generated layouts from the baselines (i.e., LayoutTransformer, BLT) and our approach, as well as the real layouts, in Figure~\ref{fig:example}. 
We can observe a typical error in the AR model LayoutTransformer that when it generates an inferior element of large area, it is impossible to revise and thus the subsequent elements can only be placed over, resulting in bad occlusion. Similar, BLT also suffers from the occlusion issue which might caused by the simple but weak heuristics that detect the erroneous tokens.
Being compared, our model generates more balanced layouts that elements are evenly distributed, which indicates that our model has a stronger capability of capturing global contexts and locating mistakes. Moreover, our generated results enjoys better alignment and less overlap in most cases.

\subsection{Model Analysis}
To better understand what our model has learned, this subsection conducts several ablation studies on different model components. The following experiments are conducted on PubLaynet. In addition, we show more results in the suppl.

\begin{figure}[htb]
\centering
\includegraphics[scale=0.40]{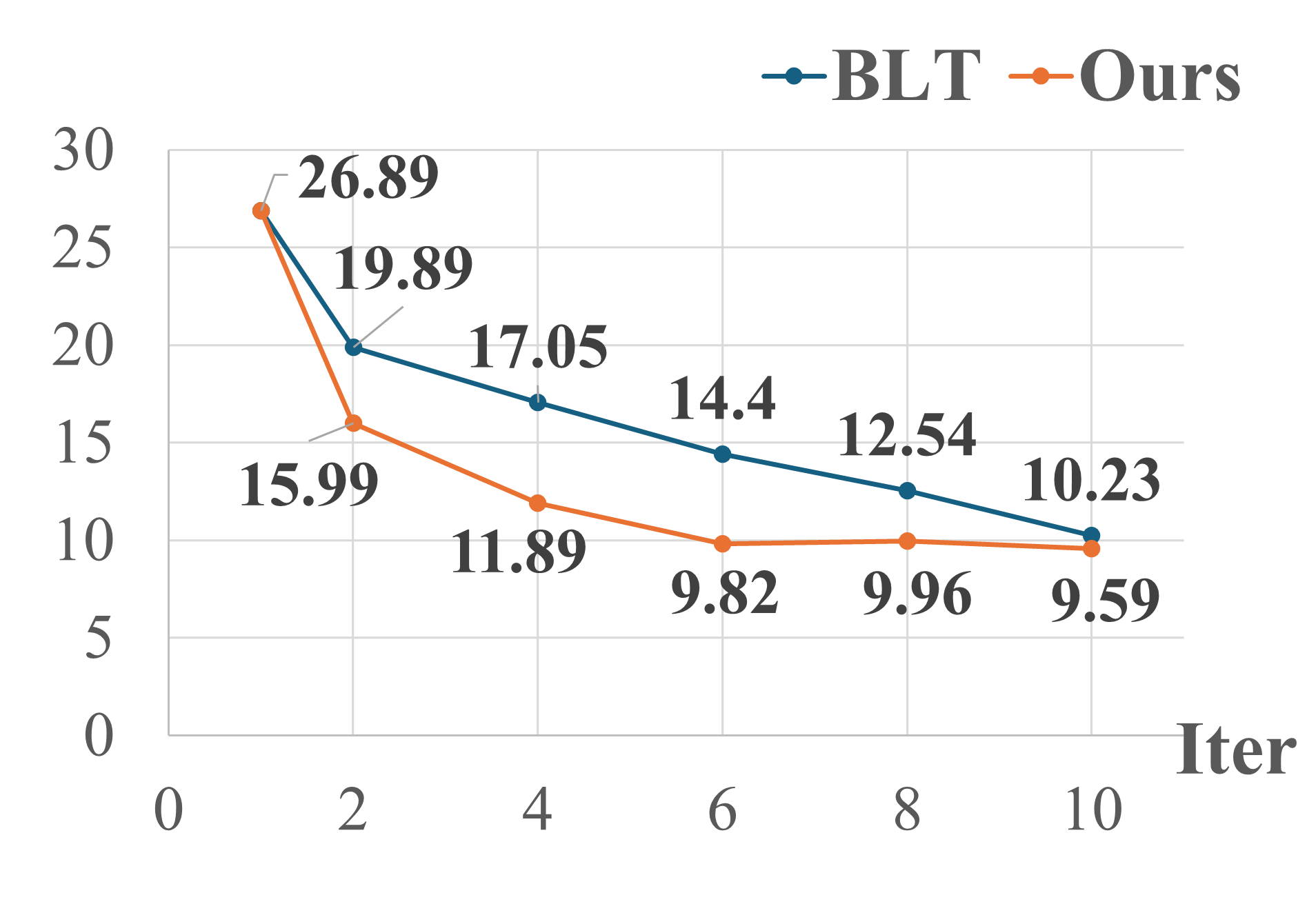}
\caption{Overlap performance of BLT and our models with different decoding iterations.}
\label{fig:performanceAlongIter}
\end{figure}

\paragraph{Performance along Iterations.} Figure~\ref{fig:performanceAlongIter} shows the overlap performance of BLT and our model in different iterations. As we can see, our model gets significantly improved from iteration 1 to 10 (smaller overlap is better), which means that the iterative decoding with our learning-based locator is effective to recognize erroneous element attributes and refine the layouts. Furthermore, our model gets faster convergence than BLT, achieving stable results in around iteration 6. As the number of iterations increases, the performance gain is becoming smaller, which is expected since the generated layouts have been refined to be better after several iterations of mask-predict.

\paragraph{Locator Ablations.} This experiment investigates the impacts of different locator settings in Table~\ref{tab:ablation-diffLocator}. Besides the end-to-end results, we show the intermediate detection accuracy of the locator. Here we obtain two interesting observations: 

\textbf{(1) pixel space is more informative than object space for capturing spatial patterns.} Similar to the previous toy experiment in Table~\ref{tab:analysis-layoutMask}, we try the two data representation for locator input. To exclude the factor of model backbone and ensure a fairer comparison, we try different locator backbones (i.e., Convolution-based Faster-RCNN and Transformer-based DETR~\cite{detr2020}) with similar parameter size. As we can see that different backbones in the pixel space perform comparably well, and are both significantly better than the one in object space. Specifically, locator in pixel space (FasterRCNN) detects the inferior elements more accurately (f1 score 59.82 compared to 29.52 in object space setting), and thus results in a better end-to-end results. This further indicates that pixel space can capture more informative spatial patterns in the layout. 

\textbf{(2) training data distribution matters for aligning the locator and the decoder.} As previously mentioned, we train the locator using the decoder generated outputs, which aims to align better with the decoder's distribution. To support our argument, we use a different training set for locator where we add random noise to the element attributes in real layouts, which is the same as the data construction in Table~\ref{tab:analysis-layoutMask}. From the last two rows in Table~\ref{tab:ablation-diffLocator}, we can see that using decoder-output data (Dec-output) is more effective than the one using random-noise data (RandNoise), indicating that the distribution alignment can help locator better focus on the decoder's error cases and result in higher performance.

\begin{table}[t]
    \centering
    \resizebox{1\columnwidth}{!}{
    \begin{tabular}{lccccc}
    \toprule
     & \multicolumn{3}{c}{\textbf{cls.}} & \multicolumn{2}{c}{\textbf{e2e}}\\
    \cmidrule(lr){2-4} \cmidrule(lr){5-6}   
    Settings & p & r & f1 & PixelFID↓  & Overlap↓ \\
    \cmidrule(lr){1-6}
    Dec-only  &  - & - & - & 113.36 & 26.89  \\
    \midrule
    Object   &  35.55 & 25.24 & 29.52 & 14.11  & 11.11 \\
    Pixel (FasterRCNN)  & \textbf{62.75} & 57.14 & \textbf{59.82} & \textbf{2.33} & \textbf{7.57} \\
    Pixel (DETR)  &  36.10 & \textbf{98.53} & 52.84 & 3.12  & 7.71\\
    \midrule
    Dec-output  & \textbf{62.75} & \textbf{57.14} & \textbf{59.82} & \textbf{2.33}  & \textbf{7.57} \\
    RandNoise  & 20.40 & 53.45  & 29.53 & 19.23   & 19.69 \\
    \bottomrule
    \end{tabular}
    }
    \caption{Performances under different locator settings, including \textbf{input layout representation} (Object, Pixel), \textbf{backbone architecture} (FasterRCNN, DETR), \textbf{training data source} (Dec-Output, RandNoise).}
    \label{tab:ablation-diffLocator}
\end{table}

\paragraph{Visualization of Decoder-Wireframe Cross-Attention.} Different from BLT, our decoder is conditioned on a wireframe image which is rendered from the decoder's output sequence in the previous iteration. To investigate what the model has captured, we visualize the cross attention weight (Equation~\ref{equ:crossAtten}) in Figure~\ref{fig:CrossAttenHeatMap}.Both the two elements' category tokens have correctly attended the corresponding regions in the wireframe, which means that the decoder can effectively utilize contexts in pixel space for better generation.

\section{Conclusion}
NAR layout generation shows competitive results compared to the AR approach. In this paper, we conduct a detailed analysis to compare the two frameworks, as well as different layout representations (i.e., object and pixel). Based on our observation, we propose a wireframe locator which explicitly learns to detect the erroneous tokens with visual input. Experiments show that our approach can achieve remarkable improvements over previous models. Further analysis reveals patterns that our model has learned. In the future, we would like to extend to content-aware generation. How to distill the knowledge in large-scale multimodal model for graphic design is another interesting direction to explore.

\bibliography{aaai24}


\end{document}




\begin{figure}[htb]
\centering
\includegraphics[scale=0.45]{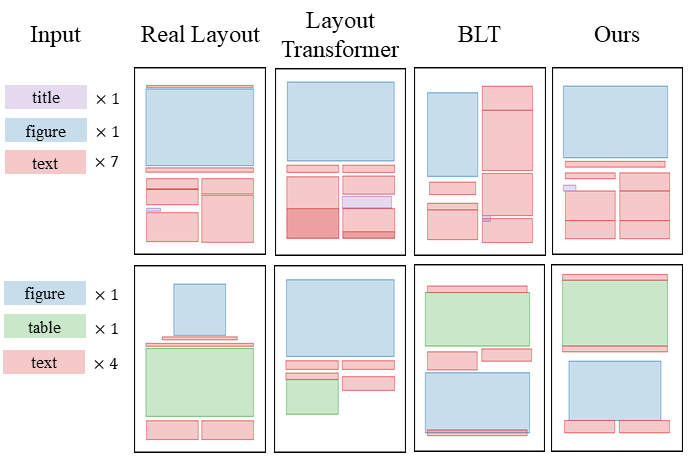}
\includegraphics[scale=0.445]{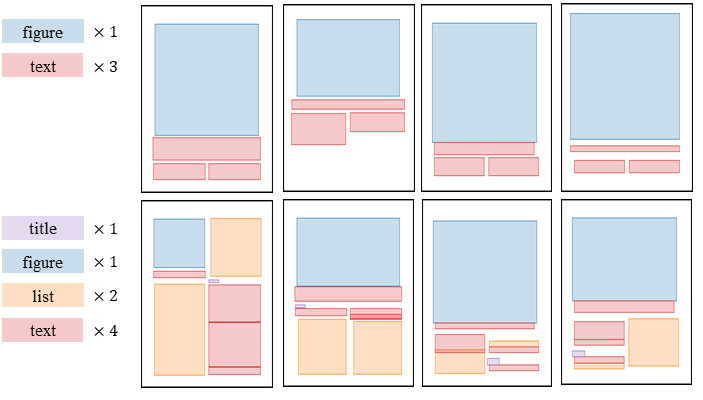}
\caption{Qualitative results on Publaynet.
}
\label{fig:example-publaynet}
\end{figure}

\begin{figure*}[htb]
\centering
\includegraphics[scale=0.43]{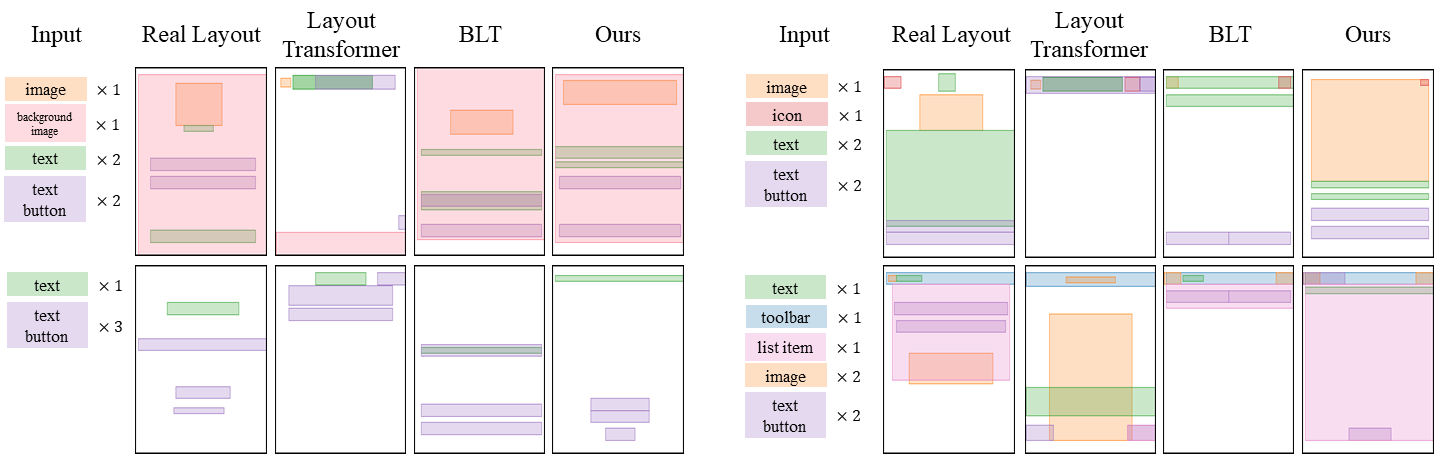}
\caption{Qualitative results on Rico.
}
\label{fig:example-rico}
\end{figure*}

\section{Implementations Details}
Our decoder is a transformer-based model with 4 layers implemented by PyTorch. The number of heads is set to 8 and the embedding dim is 512. We use a warmup schedule for the decoder and the warmup iteration is set to 4000. For our locator, we use FasterRCNN with ResNet50 and do not use pre-trained weights. While generating train data for locator, we set $\alpha_1$ to 10000, $\alpha2$ to 64, $\alpha_3$ and $\alpha_4$ to 32. We train both two models with AdamW optimizer with $\beta_1 = 0.9$ and $\beta_2=0.98$. The learning rate is set to 5e-5 for the decoder and 1e-4 for the locator. The batch size is 128 for both two models. We use the implementation from LayoutGAN++\footnote{https://github.com/ktrk115/const\_layout} for all the evaluation metrics. For SeqFID, we also adopt the released model weight of LayoutGAN++.

By the way, we have reproduced the baseline BLT and also ran its official code\footnote{https://shawnkx.github.io/blt} at the same time. Since our results are better, the baseline results in all tables are from the version we reproduced. We have placed the comparison between the official code results and our results in Table~\ref{tab:BLT-compararion}.

\begin{table}[t]
\centering
\resizebox{1\columnwidth}{!}{
\begin{tabular}{lccccc}
\toprule
\textbf{Dataset: Publaynet}  & SeqFID$\downarrow$ & PixelFID$\downarrow$ & Max IoU$\uparrow$ & Overlap$\downarrow$ & Alignment$\downarrow$ \\
\cmidrule(lr){1-6}
BLT (official) & 10.78 & 35.71 & 0.2616 & 17.38 & \textbf{0.0600} \\
BLT (our implementation) & \textbf{7.88} & \textbf{4.60} & \textbf{0.3865} & \textbf{8.73} & 0.0765 \\
\bottomrule
\toprule
\textbf{Dataset: Rico}  & SeqFID$\downarrow$ & PixelFID$\downarrow$ & Max IoU$\uparrow$ & Overlap$\downarrow$ & Alignment$\downarrow$ \\
\cmidrule(lr){1-6}
BLT (official) & \textbf{30.56} & 12.40 & 0.4716 &86.23 & 0.3060 \\
BLT (our implementation) & 32.67 & \textbf{3.80} & \textbf{0.5669} & \textbf{72.64} & \textbf{0.2361}  \\
\bottomrule
\end{tabular}
}
\caption{Conditional generation results of BLT official code and our implements when given element category (C→SP).}
\label{tab:BLT-compararion}
\vspace{-5pt}
\end{table}

\section{More Qualitative Results}
Here we show more generated layouts from the baselines (i.e., LayoutTransformer, BLT) and our approach, as well as the real layouts, in Figure~\ref{fig:example-publaynet} and Figure~\ref{fig:example-rico}.

\section{Unconditional Generation}

We further show unconditional generation results in Table~\ref{tab:unconstrain}. As we can see, utilizing only the wireframe-conditioned decoder (dec-only) already yields better performance compared to BLT (e.g., PixelFID is 28.18 versus 48.10 in BLT on Publaynet, and 2.01 versus 74.09 in BLT on Rico). By incorporating the locator, we surpass our own dec-only approach and achieve the best results in most of the metrics.

\begin{table*}[ht]
    \centering
    \setlength{\tabcolsep}{1mm}
    \resizebox{2\columnwidth}{!}{
    \begin{tabular}{lcccccccccc}
    \toprule
        \multirow{2}{*}{\textbf{Models}} & \multicolumn{5}{c}{\textbf{Publaynet}} & \multicolumn{5}{c}{\textbf{Rico}}\\
        \cmidrule(lr){2-6} \cmidrule(lr){7-11}
         & SeqFID↓ & PixelFID↓ & Max IoU↑ & Overlap↓ & Alignment↓ & SeqFID↓ & PixelFID↓ & Max IoU↑ & Overlap↓ & Alignment↓ \\
         \cmidrule(lr){1-11}
         BLT & \textbf{77.90} & 48.10 & 0.2941 & 19.03 & 0.0975 & 151.31 & 74.09 & \textbf{0.5986} & 147.88 & 0.3602 \\
         \textbf{Ours (dec-only)} & 137.77 & 28.18 & 0.5821 & 17.71 & 0.1217 & 33.49 & 2.01 & 0.5776 & 62.82 & 0.2961 \\
         \textbf{Ours (w/ locator)}& 144.59 & \textbf{20.26} & \textbf{0.7074} & \textbf{16.94} & \textbf{0.0963} &
         \textbf{32.93} & \textbf{1.65} & 0.5783	& \textbf{62.02}	& \textbf{0.2931} \\
         \cmidrule(lr){1-11}
         Real data& 4.91 & 0.02 & 0.5300 & 0.22 & 0.0400 & 4.26 & 1.16 & 0.6600 & 48.43 & 0.2000 \\
        \bottomrule
    \end{tabular}
    }
    \caption{Unconditional generation results. The best results are in \textbf{bold}.}
    \label{tab:unconstrain}
\end{table*}

\section{Matching Examples}
\begin{figure*}[htb]
\centering
\includegraphics[scale=0.46]{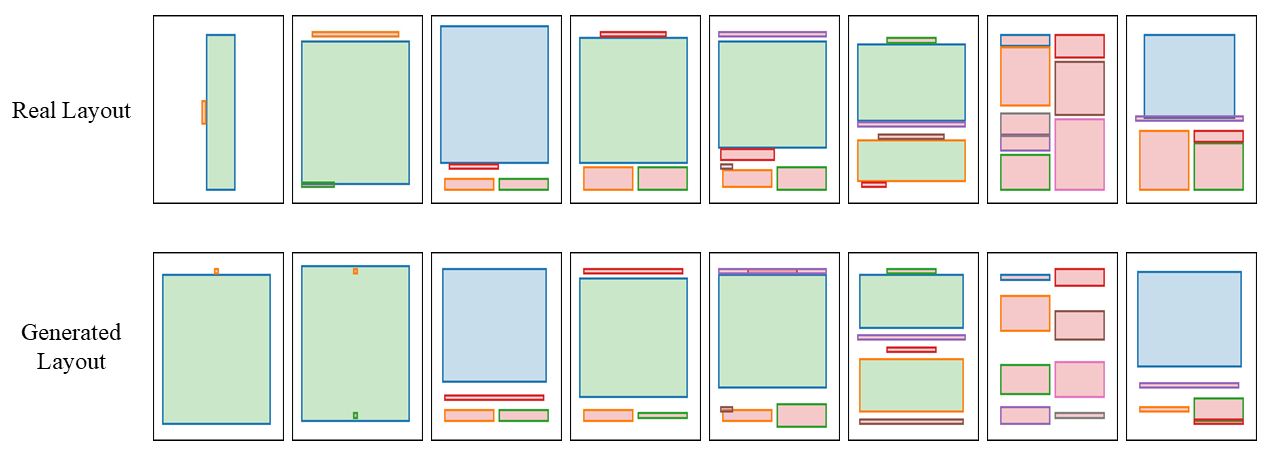}
\caption{Qualitative results of Hungarian matching in Algorithm 1 for generating train data of locator. Elements having the same color outlines in real and generated layouts are matched pairs.
}
\label{fig:example-matching}
\end{figure*}

We visualize some examples of the Hungarian matching in Algorithm 1 for generating locator train data in Figure~\ref{fig:example-matching}. As we can see, the method successfully matches elements belonging to the same category, as well as those with more similar sizes and closer positions.

